\documentclass{article}
\usepackage{iclr2027_conference,times}
\usepackage{float}
\usepackage{hyperref}
\usepackage{url}
\usepackage{booktabs}
\usepackage{subcaption}
\usepackage{makecell}
\usepackage{amsthm}
\usepackage{algorithm}
\usepackage{algorithmicx}
\usepackage{algpseudocode}
\usepackage{graphicx}
\usepackage[table]{xcolor}

\usepackage{amsmath,amsfonts,bm}

\def\eqref#1{equation~\ref{#1}}

\def\1{\bm{1}}

\DeclareMathAlphabet{\mathsfit}{\encodingdefault}{\sfdefault}{m}{sl}
\SetMathAlphabet{\mathsfit}{bold}{\encodingdefault}{\sfdefault}{bx}{n}

\newtheorem{proposition}{Proposition}
\newtheorem{lemma}{Lemma}
\newtheorem{corollary}{Corollary}

\definecolor{lightgray}{gray}{0.9}
\newcommand{\topone}[1]{\cellcolor{lightgray}#1}
\newcommand{\toptwo}[1]{\textbf{#1}}

\title{Rethinking Feature Perturbation in Single-Source Domain Generalization: A Representation-Space Analysis and a Spectral Case Study}

\author{Badr-Eddine Benelmostafa \\
  EIGSI Casablanca, Morocco\\
  \texttt{be.benelmostafa@eigsi.org}
  \And
  Sohaib Baroud \\
  EIGSI Casablanca, Morocco\\
  \texttt{sohaib.baroud@eigsi.org}
}
\iclrfinalcopy

\begin{document}
\maketitle
\lhead{} 
\renewcommand{\headrulewidth}{0pt} 
\begin{abstract}
Feature perturbation is a common strategy for single-source domain generalization (SSDG), but it is usually judged by its final task performance rather than by the representation changes it induces. We explore a different perspective: can feature perturbations be compared through a small set of measurable descriptors that describe how a representation moves? We introduce a four-dimensional representation-space descriptor, comprising feature displacement $D_F$, relational displacement $D_R$, a task-region ratio $\Theta_{\mathrm{Task}}$, and subspace deviation $\Theta_{\mathrm{Subspace}}$. We use this space as an empirical analysis lens for several existing perturbation mechanisms rather than as a presumed universal geometry. Across two object-detection settings with style and weather shifts, we observe recurring differences in the geometric profiles of perturbations and a concentration of relatively stronger OOD-performing configurations in a bounded region of the descriptor space. These observations do not establish a universal optimum, but they suggest a concrete direction for future study: whether SSDG performance may be associated with reproducible regions of representation space that can be identified without target-domain access. As a case study, we introduce Dual-Invariant Spectral Perturbation (DISP), a non-learned Fourier-phase perturbation with explicit controls over phase magnitude $\rho$, channel coupling $\kappa$, and frequency support $\mathcal{M}$. We establish energy preservation and relational invariance at full channel coupling, and empirically examine how the controls alter the proposed descriptors. DISP obtains strong OOD performance in the reported experiments, but the main purpose of the method is illustrative: it shows how explicit perturbation controls can be studied through a representation-space description. We hope this viewpoint can contribute to more interpretable, controllable, and ultimately testable approaches to feature perturbation in SSDG.
\end{abstract}
\section{Introduction}
\label{sec:intro}

Deep networks can perform remarkably well when training and test data follow similar distributions, yet deployment often violates this assumption. Changes in appearance, sensing conditions, scene composition, or weather can substantially alter the visual distribution, as encountered in UAV inspection and other real-world vision applications \citep{benelmostafa_yolov8-erfd-ap_2025,fan_robust_2025,oh_fiesta_2024}. This motivates \emph{domain generalization} (DG), which seeks models that transfer to unseen target domains without target data during training \citep{gulrajani_search_2020}. The single-source setting (SSDG) is particularly challenging because only one training environment is observed, forcing the model to construct useful variability without directly observing cross-domain representation changes \citep{qiao_uncertainty-guided_2021}. In object detection, this challenge is further complicated by appearance, texture, scale, background context, localization, and partial visibility.

Among existing approaches, feature perturbation provides a direct way to introduce variability inside the network: an intermediate representation $F$ is replaced during training by

$$
F'=\mathcal{P}(F),
$$

without requiring target-domain observations. However, perturbations are typically evaluated through downstream performance, commonly $mAP_{50}$, which does not reveal what they change internally. Two perturbations can achieve similar detection scores while differing in feature displacement, relational structure, or spatial allocation; conversely, a larger feature change does not necessarily yield better OOD generalization.

We therefore ask:
\begin{quote}
\emph{Can feature perturbations be described in a common representation-space language that is informative beyond final detection accuracy?}
\end{quote}

To explore this question, we construct a four-dimensional empirical descriptor space capturing \emph{how far} the representation moves, \emph{how much} its channel relations change, \emph{where} the displacement occurs relative to object regions, and \emph{whether} it leaves a dominant clean-feature subspace. We use these descriptors to compare several perturbation mechanisms under a common protocol. Across two benchmarks, relatively stronger OOD configurations tend to occupy a bounded portion of this space. We treat this as a hypothesis rather than a universal optimum, motivating a broader question: \emph{does SSDG exhibit a reproducible region where source diversification and preservation of task-relevant structure are jointly achieved, and can it be identified using only source-domain information?}

To make this question concrete, we introduce \emph{Dual-Invariant Spectral Perturbation} (DISP) as a simple case study. DISP perturbs Fourier phase while preserving spectral amplitude, with three explicit intervention variables: phase magnitude $\rho$, channel coupling $\kappa$, and spectral support $\mathcal{M}$. Rather than assuming that these controls directly determine the proposed descriptors, we measure how representation geometry responds to controlled changes in the operator, providing a simple setting in which perturbation design and representation analysis can be studied together.

Our contributions are intentionally exploratory:
\textbf{(i)} a four-dimensional descriptor space for comparing feature perturbations;
\textbf{(ii)} an empirical characterization of existing mechanisms and recurring descriptor patterns associated with stronger OOD performance across two benchmarks;
\textbf{(iii)} DISP as a non-learned spectral case study with explicit, measurable intervention variables; and
\textbf{(iv)} open research questions on whether descriptor spaces can become predictive, source-selectable, and useful for interpretable SSDG perturbation design.

\section{Related Work}
\label{sec:related_work}

Feature perturbation methods create training variability by modifying intermediate representations, but they differ substantially in what is changed and how the perturbation is generated. We organize the literature around these mechanisms while keeping the central question of this work in view: how can the resulting representation change be described in a common language?

\paragraph{Stochastic and statistical perturbation.}
DSU \citep{li_uncertainty_2021} models channel-wise feature statistics as random variables. Normalization Perturbation (NP) \citep{fan_normalization_2022} perturbs low-level statistics, with a particular emphasis on shallow CNN representations. MixStyle \citep{zhou_mixstyle_2024} mixes statistics across samples, while CrossNorm exchanges them \citep{tang_crossnorm_2021}. Adversarial Style Augmentation (ASA) searches for challenging perturbations in statistic space \citep{zhang_adversarial_2023}. These methods are attractive because their interventions are simple to implement, but their effect on the full representation is typically summarized through the statistics being manipulated rather than through a common representation-space description.

\paragraph{Distribution-level and cross-sample perturbation.}
EFDM extends statistical matching by transforming empirical feature distributions \citep{zhang_exact_2022}. CCFP introduces cross-sample feature perturbation under semantic consistency constraints \citep{li_cross_2023}, while Domain Feature Perturbation exploits domain information through an auxiliary classifier \citep{wang_domain_2024}. Other SSDG work has explored combined image- and feature-level perturbations, such as CPerb and MixPatch \citep{zhao_novel_2024}. Recent object-detection studies also emphasize the balance between diversity and invariance under single-source shifts \citep{danish_improving_2024,he_single-domain_2026}. These approaches broaden the space of perturbation mechanisms, but they also make direct comparison of the induced feature changes more difficult.

\paragraph{Spectral perturbation.}
Fourier methods provide a different way of constructing domain variation. AmpMix perturbs Fourier amplitude while preserving phase \citep{xu_fourier-based_2023}, motivated by the different roles of amplitude and phase in signal representation \citep{oppenheim_importance_1981}. FIESTA introduces uncertainty-guided spectral manipulation \citep{oh_fiesta_2024}, while UniFreqSDG learns spectral perturbation ranges and strengths for SSDG \citep{liu_universal_2024}. More recently, AdvSP uses learnable spectral perturbations for single-domain generalized object detection \citep{wang_adversarial_2026}. These studies support the usefulness of frequency-domain interventions, but they do not provide a shared descriptor language for comparing the representation changes produced by otherwise different mechanisms.

\paragraph{Position of this work.}

Our objective is therefore not to claim that existing perturbation methods lack control or interpretation. Several expose useful hyperparameters, and learned methods may have considerably more expressive internal control than their user-facing parameterization suggests. Instead, we ask whether their effects can be described under the same representation-space descriptors. We use this perspective to analyze existing mechanisms and then to study DISP as a simple, explicit intervention whose controls can be related to measurable representation changes.

\begin{table}[htbp]
\centering
\small
\caption{Exposed control levers of feature perturbation mechanisms.
\emph{Learned} indicates whether the perturbation itself is optimized
jointly with the task. \emph{Control levers} lists parameters exposed to
the practitioner beyond a fixed application probability; most methods
expose none, relying instead on a hardcoded scale, no free parameter, or
values predicted internally.}
\label{tab:related_work_overview}
\begin{tabular}{@{}lllc@{}}
\toprule
Method & Modified object & Learned & Control levers \\
\midrule
MixStyle \citep{zhou_mixstyle_2024} - 2024    & channel mean / std        & No  & $\alpha$ (mixing distribution) \\
CrossNorm \citep{tang_crossnorm_2021} - ICCV 2021   & channel mean / std        & No  & -- \\
DSU \citep{li_uncertainty_2021} - ICLR 2022         & channel mean / std        & No  & none (hardcoded scale) \\
NP \citep{fan_normalization_2022} - ICLR 2023      & channel mean / std        & No  & none (hardcoded scale) \\
EFDM \citep{zhang_exact_2022} - CVPR 2022          & empirical distribution    & No  & none (parameter-free) \\
AmpMix \citep{xu_fourier-based_2023} - 2023   & Fourier amplitude         & No  & $\alpha$ (mixing bound) \\
FIESTA \citep{oh_fiesta_2024} - 2024      & Fourier amplitude         & No  & noise scale \\
CCFP \citep{li_cross_2023} - ICCV 2023            & cross-sample stat.\ shift & \textbf{Yes} & none (architecture only) \\
UniFreqSDG \citep{liu_universal_2024} - ACM MM 2024   & amplitude, radius, gating & \textbf{Yes} & none (predicted internally) \\
AdvSP \citep{wang_adversarial_2026} - 2026    & Fourier spectrum          & \textbf{Yes} & none (predicted internally) \\
\midrule
\textbf{DISP (ours)}                    & Fourier phase             & No  & $\rho$, $\kappa$, band $\mathcal{M}$ \\
\bottomrule
\end{tabular}
\end{table}

\section{A Representation-Space Descriptor View}
\label{sec:framework}

\subsection{Motivation}

A perturbation is usually defined by the operation it applies: changing statistics, matching a distribution, exchanging information across samples, or modifying a spectrum. Such definitions describe the mechanism, but they do not necessarily reveal the resulting representation shift in a comparable form. We therefore introduce a descriptor view based on four complementary questions:
\emph{how far} does the representation move,
\emph{how much} does its relational structure change,
\emph{where} does the displacement occur with respect to object regions, and
\emph{whether} does the displacement leave a dominant clean-feature subspace?
The resulting descriptor is
\[
\mathcal{G}(F,F')=
\left(
D_F,
D_R,
\Theta_{\mathrm{Task}},
\Theta_{\mathrm{Subspace}}
\right).
\]

We emphasize that this construction is an \emph{analysis lens}, not a claim that these four quantities form a complete or canonical geometry of representation space. Each descriptor captures one aspect of the perturbation, and each also depends on design choices that we make explicit below.

\subsection{Descriptors}

Let $F\in\mathbb{R}^{C\times H\times W}$ denote a clean feature map, $F'=\mathcal{P}(F)$ its perturbed counterpart, and $\Delta=F'-F$. Let $F_{(2)},\Delta_{(2)}\in\mathbb{R}^{C\times HW}$ denote their spatially flattened forms, and let $\varepsilon>0$ denote a numerical stabilizer.

\paragraph{Feature displacement $D_F$.}
\[
D_F=\frac{\|\Delta\|_F}{\|F\|_F+\varepsilon}.
\]
$D_F$ measures the relative size of the induced feature change. It is intentionally simple: the purpose is to provide a common magnitude reference rather than to assert that larger displacement is better or worse.

\paragraph{Relational displacement $D_R$.}
We define the channel Gram matrix
\[
R(F)=\frac{1}{HW}F_{(2)}F_{(2)}^{\top},
\]
and measure its change as
\[
D_R=\frac{\|R(F')-R(F)\|_F}{\|R(F)\|_F+\varepsilon}.
\]
This descriptor captures changes in second-order inter-channel structure and complements $D_F$: two perturbations can have similar displacement magnitude while modifying channel relations differently.

\paragraph{Task-region ratio $\Theta_{\mathrm{Task}}$.}
Let $M\in\{0,1\}^{H\times W}$ denote the union of projected ground-truth object regions on the feature grid and let $\bar M=1-M$. For a spatial mask $m$, define
\[
D(\Delta;m)=
\frac{\|\Delta\odot m\|_F}
{\|F\odot m\|_F+\varepsilon}.
\]
We then use
\[
\Theta_{\mathrm{Task}}
=
\frac{D(\Delta;\bar M)}{D(\Delta;M)+\varepsilon}.
\]
Values above one indicate proportionally larger displacement in background than in object regions. This quantity should be interpreted as a spatial allocation descriptor, not as a direct measure of task importance.

\paragraph{Subspace deviation $\Theta_{\mathrm{Subspace}}$.}
Let $V_K$ contain the top-$K$ right singular vectors of the clean $F_{(2)}^{\top}$, with $K=\lceil k_{\mathrm{ratio}}C\rceil$. We define
\[
E_{\mathrm{tot}}=\|\Delta_{(2)}\|_F^2,
\qquad
E_{\parallel}=\|\Delta_{(2)}^{\top}V_K\|_F^2,
\]
and
\[
\Theta_{\mathrm{Subspace}}
=\frac{E_{\mathrm{tot}}-E_{\parallel}}{E_{\mathrm{tot}}}
\in[0,1].
\]
Larger values indicate that a larger fraction of perturbation energy lies outside the chosen dominant clean-feature subspace. The reference subspace is always computed from $F$, not $F'$, providing a common frame across methods. As with all descriptors, the choice of $k_{\mathrm{ratio}}$ is a modeling choice; we use $0.2$ throughout the present analysis.

\subsection{What This Descriptor View Can and Cannot Establish}

The descriptor view supports comparisons that are difficult to make from the perturbation operator alone. It allows us to ask whether methods with similar downstream performance induce similar representation changes, whether large changes are systematically associated with OOD performance, and whether object-region allocation or subspace deviation provides information not captured by a single magnitude measure. At the same time, the descriptors do not establish causality. A correlation between a descriptor and OOD performance can arise from the architecture, training dynamics, data composition, or interactions between the perturbation and the learned representation. Nor do we assume that the favorable region identified in our current experiments is universal. The intended contribution is to formulate a measurable object for future study: a representation descriptor space in which perturbations can be compared, analyzed, and eventually, perhaps, selected without target-domain supervision.

\section{DISP: A Spectral Case Study}
\label{sec:method}

\subsection{From Analysis to an Explicit Intervention}

The descriptor view naturally suggests a second question: if perturbation effects can be measured in a common representation space, can we design an intervention whose mechanism exposes interpretable variables that can be systematically varied?

We study this question through \emph{Dual-Invariant Spectral Perturbation} (DISP). DISP is not proposed as a solution to the descriptor-selection problem. Instead, it is a deliberately simple case study that makes the interaction between perturbation controls and representation descriptors explicit.

\subsection{DISP Operator}

Let $F\in\mathbb{R}^{C\times H\times W}$ be an intermediate feature map and $X=\mathcal{F}_2(F)$ its per-channel spatial Fourier transform,
\[
X_c(u,v)=A_c(u,v)e^{i\varphi_c(u,v)}.
\]
DISP preserves amplitude and perturbs phase within a selected band $\mathcal{M}$:
\[
\varphi'_c(u,v)=
\varphi_c(u,v)+\rho M_{\mathcal M}(u,v)\delta_c(u,v),
\]
\[
F'=\mathcal{F}_2^{-1}\!\left(A\odot e^{i\varphi'}\right),
\qquad
\delta_c=
\kappa z^{\mathrm{shared}}+
\sqrt{1-\kappa^2}z_c^{\mathrm{indep}},
\]
where the shared field is common to channels and the independent field is sampled per channel, both from standard Gaussian noise. The perturbation is normalized to unit RMS in the selected band before being scaled by $\rho$, and the DC component is excluded.

The controls have distinct \emph{mechanistic} roles:
$\rho\ge0$ sets the RMS magnitude of the phase perturbation,
$\kappa\in[0,1]$ determines the balance between shared and channel-specific phase variation, and $\mathcal{M}$ specifies the spectral support of the perturbation. Importantly, we do not assume that these variables independently determine the four descriptors. Their effect on representation geometry is measured empirically.

\subsection{Analytical Properties}

\paragraph{Energy preservation.}
Because DISP modifies phase while preserving spectral amplitude, under an orthonormal Fourier transform,
\[
\|F'\|_F=\|F\|_F.
\]
Consequently,
\[
0\le D_F\le2,
\qquad
\rho\rightarrow0\Longrightarrow D_F\rightarrow0.
\]
The first statement follows from equal feature energy and the triangle inequality; the second follows from the continuity of the inverse Fourier transform.

\paragraph{Relational invariance at full coupling.}
For the channel Gram matrix above, Parseval's identity gives
\[
R(F)_{cc'}=
\frac{1}{HW}\sum_{u,v}
A_cA_{c'}\cos(\varphi_c-\varphi_{c'}).
\]
When $\kappa=1$, every channel receives the same phase offset, so the relative phase between channel pairs is unchanged. Under the corresponding full-spectrum/Hermitian reconstruction,
\[
\kappa=1
\Longrightarrow
R(F')=R(F)
\Longrightarrow
D_R=0.
\]
For intermediate $\kappa$, we make no monotonicity claim: the resulting relational change depends on the existing phase structure and the sampled channel-specific component.

\paragraph{Frequency localization.}
The band $\mathcal{M}$ specifies where the phase perturbation is applied. Fourier analysis motivates treating frequency localization as a meaningful intervention variable because different frequency components can contribute to different spatial scales \citep{oppenheim_importance_1981}. We do not derive a deterministic mapping from low-, mid-, or high-frequency perturbations to $\Theta_{\mathrm{Task}}$ or $\Theta_{\mathrm{Subspace}}$; those effects are treated as empirical questions.

\section{Experiments and Analysis}
\label{sec:experiments}

\subsection{Experimental Setup}
\label{sec:exp_setup}

We use a fixed YOLO26-nano detector implemented in Ultralytics~v8.4.49. All methods use the same detector, optimization protocol, image preprocessing, perturbation probability, and perturbation locations; only the perturbation mechanism changes. Motivated by the observation that low-level domain-sensitive statistics are strongly represented in shallow CNN layers \citep{fan_normalization_2022}, we insert perturbation modules at backbone layers $0$, $1$, and $3$ (the $P3/8$ transition).  We evaluate two complementary settings: Pascal VOC~2007 as the source with unseen Clipart, Watercolor, and COCO targets, and clear-weather IDID as the source with unseen rain, fog, and snow conditions. Dataset composition and splits are given in Appendix~\ref{app:datasets}. We compare source-only training (Baseline) with CCFP, AmpMix, MixStyle, DSU, EFDM, FIESTA, UniFreqSDG, NP$_{\mathrm{Original}}$, and DISP. OOD performance is reported using $\mathrm{mAP}_{50}$ with perturbations disabled during validation and inference. Geometric descriptors are computed on paired clean and perturbed source-domain evaluation features at $P3/8$. 

\subsection{Main Out-of-Domain Detection Results}
\label{sec:main_results}

\begin{figure}[h]
    \centering
    \includegraphics[width=\linewidth]{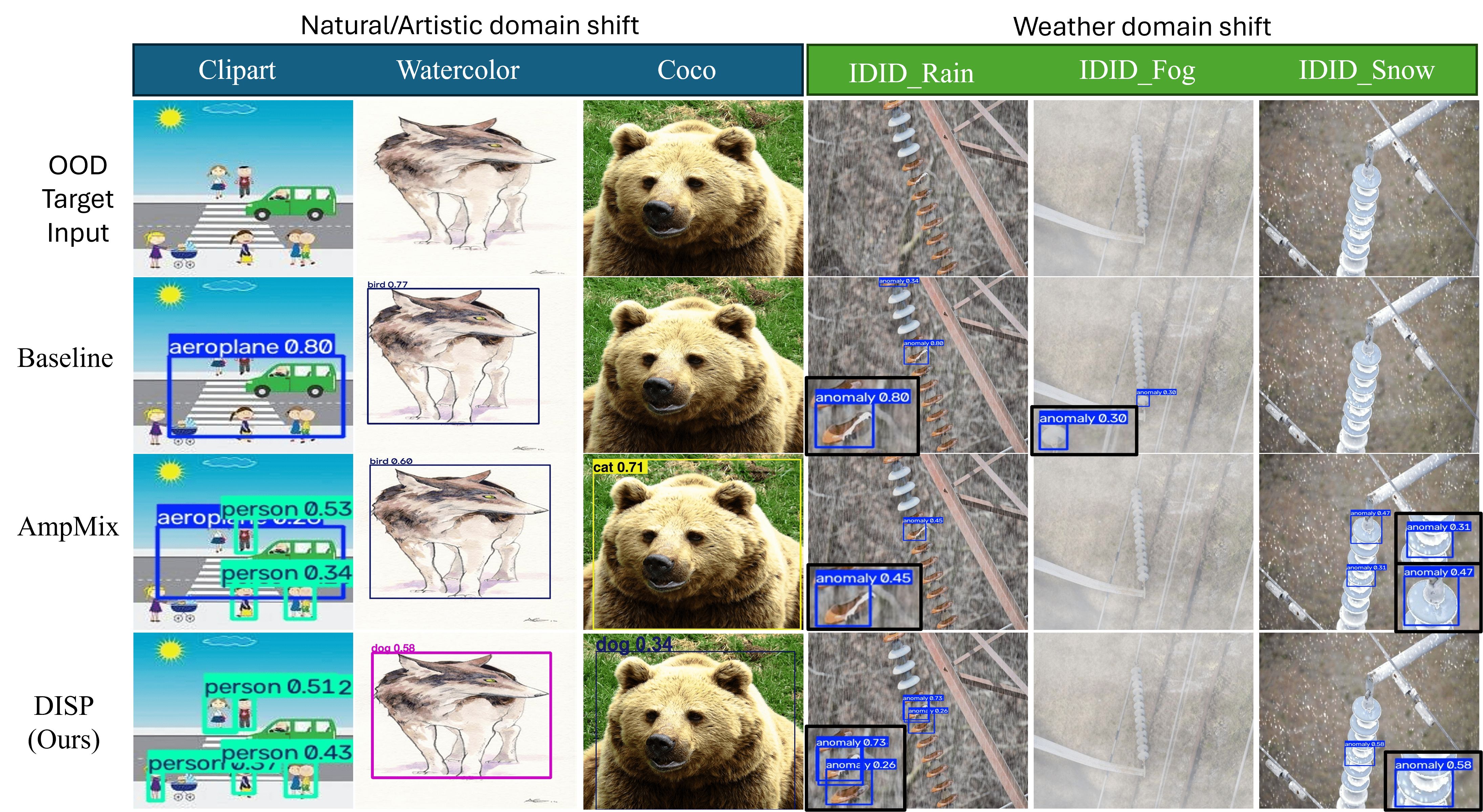}
    \caption{Representative detections on unseen style and weather domains. The examples illustrate different responses of perturbation mechanisms under domain shift; they are intended as qualitative context rather than quantitative evidence.}
    \label{fig:qualitative}
\end{figure}

Table~\ref{tab:main_merged_with_source} reports the source and unseen-target $\mathrm{mAP}_{50}$ values obtained in our current runs. In these experiments, DISP has the highest reported mean OOD score among the compared configurations on both settings: $9.00\%$ on the VOC family and $48.70\%$ on IDID. The absolute margins over AmpMix are small ($+0.47$ and $+0.33$ points, respectively), so we do not interpret them as evidence of a large performance improvement. Rather, they show that the proposed spectral intervention is at least competitive under the present protocol. On VOC, DISP obtains the highest score among the compared methods on Clipart and Watercolor, while other methods remain competitive on COCO. On IDID, DISP obtains the highest reported Rain score and the highest source-domain score in our table, while DSU and FIESTA are stronger on Snow and Fog, respectively. This variation is useful for our main question because it illustrates that no single perturbation mechanism dominates all target conditions. Figure~\ref{fig:qualitative} shows representative detections. On Clipart, DISP recovers additional persons in one example where Baseline and AmpMix produce an incorrect \emph{aeroplane} prediction. In other style-shift examples, competing methods give different responses to objects outside the benchmark label space; we interpret these cases only as evidence of different model responses, not as proof that one prediction is semantically correct. The fog example similarly illustrates that a perturbation useful under one weather condition need not preserve every anomaly detection under another condition. These observations motivate examining the representation-level behavior rather than relying on isolated qualitative outcomes.

\begin{table}[H]
\centering
\caption{Single-source domain generalization results on Pascal VOC 2007
and IDID. Source performance (VOC$_{\mathrm{Orig}}$ and IDID Orig.) is
evaluated on the training domain. Mean OOD target performance is computed
over the unseen target domains only. \topone{Light gray} denotes the best
method (Top 1); \toptwo{Bold} denotes the second-best (Top 2).}
\label{tab:main_merged_with_source}
\resizebox{\textwidth}{!}{
\begin{tabular}{l|ccccc|ccccc}
\toprule
& \multicolumn{5}{c|}{\textbf{Benchmark 1}} &
\multicolumn{5}{c}{\textbf{Benchmark 2}} \\
\textbf{Method}
& VOC$_{\mathrm{Orig}}$ & Clipart & Watercolor & COCO & Mean OOD
& IDID$_{\mathrm{Orig}}$ & IDID$_{\mathrm{Snow}}$ & IDID$_{\mathrm{Rain}}$ & IDID$_{\mathrm{Fog}}$ & Mean OOD \\
\midrule
Baseline
& 25.4 & 4.6 & 4.7 & 9.3 & 6.20
& \toptwo{71.2} & 43.3 & 52.5 & \toptwo{45.4} & 47.07 \\
DSU (ICLR 2022)
& 25.0 & 4.7 & 8.7 & 9.3 & 7.57
& 68.1 & \topone{47.1} & 49.8 & 40.7 & 45.87 \\
NP$_{\mathrm{Original}}$ (ICLR 2023)
& 25.2 & 5.1 & 6.5 & 9.0 & 6.87
& 68.8 & 45.7 & 47.9 & 40.4 & 44.67 \\
EFDM (CVPR 2022)
& 24.0 & 5.7 & 8.1 & 8.4 & 7.40
& 62.9 & 40.7 & 50.2 & 42.4 & 44.43 \\
FIESTA
& 26.5 & 6.1 & 7.7 & 9.5 & 7.77
& \toptwo{71.2} & 40.5 & 51.9 & \topone{45.9} & 46.10 \\
MixStyle
& 25.7 & \topone{6.9} & 7.4 & 9.6 & 7.97
& 64.7 & 44.5 & 48.7 & 43.2 & 45.47 \\
UniFreqSDG (ACM MM 2024)
& 26.9 & 6.1 & 8.5 & \topone{9.9} & 8.17
& 67.4 & 44.4 & \toptwo{54.2} & 38.7 & 45.77 \\
CCFP (ICCV 2023)
& \toptwo{27.6} & 4.8 & 9.2 & \topone{9.9} & 7.97
& 67.1 & 42.6 & 45.8 & 38.6 & 42.33 \\
AmpMix
& \topone{27.8} & 6.1 & \toptwo{9.8} & \toptwo{9.7} & \toptwo{8.53}
& 69.8 & \toptwo{46.3} & 53.4 & \toptwo{45.4} & \toptwo{48.37} \\
\textbf{DISP (Ours)}
& \topone{27.8} & \toptwo{6.7} & \topone{10.7} & 9.6 & \topone{9.00}
& \topone{72.2} & 45.1 & \topone{55.8} & 45.2 & \topone{48.70} \\
\bottomrule
\end{tabular}}
\end{table}

\subsection{Descriptor Space and OOD Performance}
\label{sec:geometry}

The preceding table establishes the task-level outcomes. We next ask whether the methods also differ systematically in the representation descriptors introduced in Section~\ref{sec:framework}. We profile the same mechanisms at $P3/8$ using paired clean and perturbed features.

Table~\ref{tab:geometric_comparison} shows that the perturbation mechanisms occupy distinct regions of the descriptor space. EFDM is a clear outlier in the current measurements, with the largest $D_F$, $D_R$, and $\Theta_{\mathrm{Task}}$ on both benchmarks. DSU has the largest reported $\Theta_{\mathrm{Subspace}}$ on both settings. These observations by themselves say nothing about whether the corresponding coordinates are good or bad; they simply show that different perturbation rules induce measurably different representation changes.

\begin{figure}[h]
    \centering
    \includegraphics[width=\linewidth]{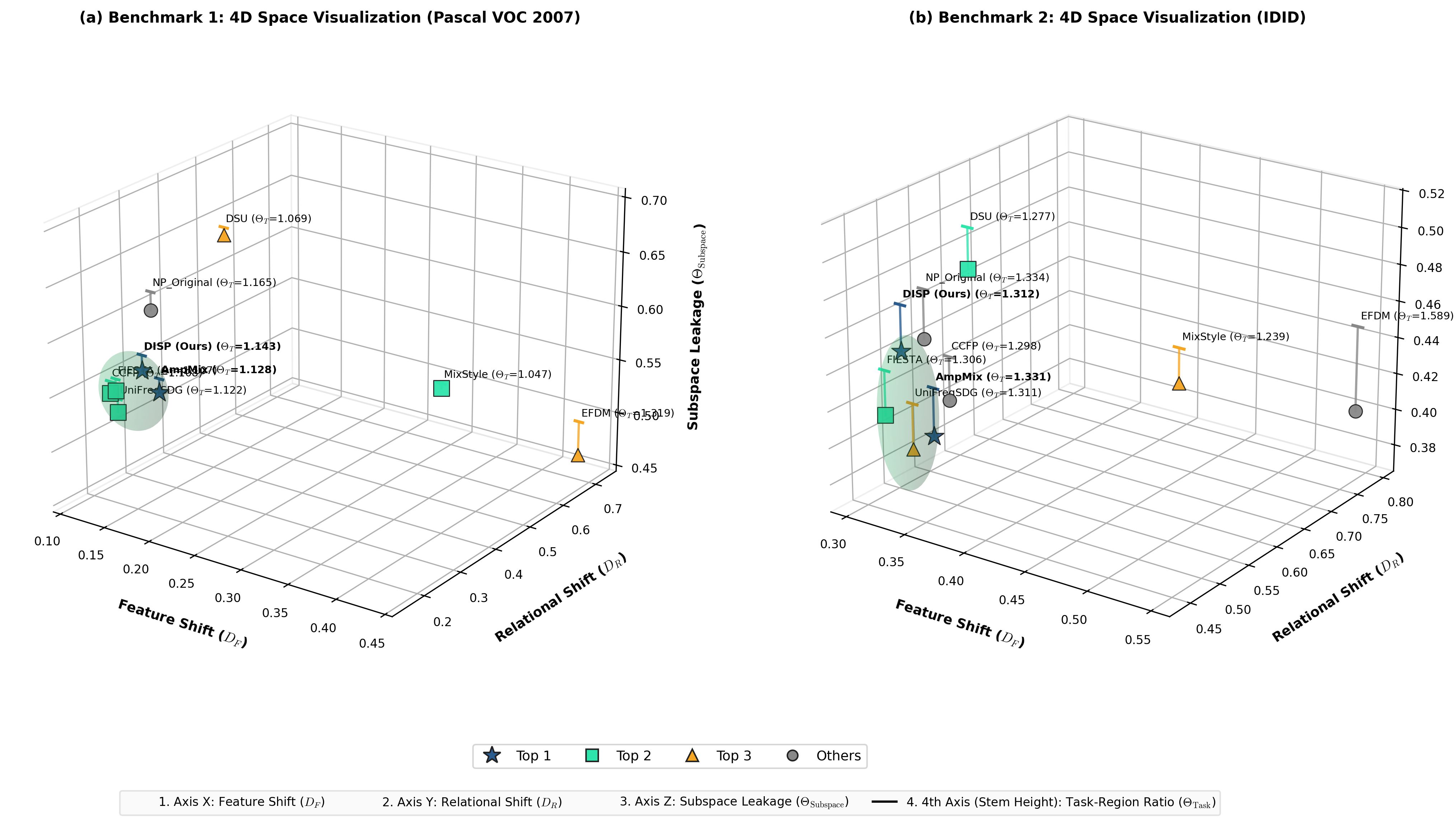}
    \caption{Proposed descriptor space on (a) Pascal VOC and (b) IDID. The shaded area is a visual guide highlighting a bounded region in which several relatively strong OOD-performing configurations occur. It is not an analytically defined optimum and should be interpreted as an empirical observation to be tested in broader settings.}
    \label{fig:sweetspot}
\end{figure}

More interestingly, several of the methods with relatively strong OOD performance appear within a bounded portion of the four-dimensional descriptor space. Figure~\ref{fig:sweetspot} visualizes this pattern. We call the region a \emph{favorable region} only in the descriptive sense that the currently plotted higher-performing configurations tend to occur there. We do not define a universal boundary, fit a predictive classifier, or claim that the region is optimal. The appropriate interpretation is therefore a research hypothesis: perhaps SSDG perturbations that generalize well do not need to maximize any individual descriptor, but instead occupy a constrained combination of representation displacement, relational change, task-region allocation, and subspace exploration.

This observation is deliberately weaker than a geometric law. With only two benchmarks, one architecture, and a small set of perturbation mechanisms, the current evidence cannot establish that the same region would persist under different backbones, tasks, or shifts. What the experiment does provide is a concrete empirical object that can be investigated further.

\begin{table}[H]
\centering
\scriptsize
\setlength{\tabcolsep}{3pt}
\caption{Representation descriptors measured at $P3/8$ on the source-domain evaluation split. Values are means over valid samples under the protocol in Appendix~\ref{app:geometry}.}
\label{tab:geometric_comparison}
\begin{tabular}{lcccc|cccc}
\toprule
\textbf{Method}
& \multicolumn{4}{c|}{\textbf{Pascal VOC 2007}}
& \multicolumn{4}{c}{\textbf{IDID}} \\
& $D_F$ & $D_R$ & $\Theta_{\mathrm{Task}}$ & $\Theta_{\mathrm{Subspace}}$
& $D_F$ & $D_R$ & $\Theta_{\mathrm{Task}}$ & $\Theta_{\mathrm{Subspace}}$ \\
\midrule
AmpMix
& 0.165 & 0.227 & 1.128 & 0.551
& 0.338 & 0.488 & 1.331 & 0.403 \\
CCFP
& 0.127 & 0.188 & 1.108 & 0.549
& 0.344 & 0.503 & 1.298 & 0.421 \\
FIESTA
& 0.132 & 0.191 & 1.107 & 0.552
& 0.312 & 0.459 & 1.306 & 0.415 \\
NP$_{\mathrm{Original}}$
& 0.168 & 0.203 & 1.165 & 0.631
& 0.337 & 0.477 & 1.334 & 0.458 \\
UniFreqSDG
& 0.133 & 0.193 & 1.122 & 0.532
& 0.330 & 0.469 & 1.311 & 0.398 \\
DSU
& 0.197 & 0.325 & 1.069 & 0.684
& 0.348 & 0.528 & 1.277 & 0.489 \\
EFDM
& 0.430 & 0.723 & 1.319 & 0.461
& 0.543 & 0.795 & 1.589 & 0.400 \\
MixStyle
& 0.358 & 0.511 & 1.047 & 0.547
& 0.468 & 0.636 & 1.239 & 0.430 \\
\textbf{DISP (Ours)}
& 0.159 & 0.198 & 1.143 & 0.575
& 0.325 & 0.463 & 1.312 & 0.452 \\
\bottomrule
\end{tabular}
\end{table}

\subsection{Controlled Exploration with DISP}
\label{sec:controlled_displacement}

We next use DISP to examine whether explicit perturbation variables produce systematic trajectories in the descriptor space. We sweep phase magnitude $\rho\in\{0.1,0.5,0.9\}$, channel coupling $\kappa\in\{0.1,0.5,0.9\}$, and frequency support $\mathcal{M}\in\{\text{low},\text{mid},\text{high},\text{all}\}$. For comparison, AmpMix is evaluated over $\alpha\in\{0.1,0.3,0.6,0.9\}$. The full values are reported in Appendix~\ref{app:navigation}. Figure~\ref{fig:navigation} shows that varying $\rho$ produces the clearest progression in feature displacement: in the mid-band sweep, larger $\rho$ substantially increases $D_F$. Changes in $\kappa$ and $\mathcal{M}$ affect combinations of relational and directional descriptors, but not necessarily in a monotonic one-coordinate fashion. This is consistent with the distinction made in Section~\ref{sec:method}: the parameters have interpretable meanings at the operator level, while their representation-level consequences emerge from the interaction with the learned features. The reported VOC sweep also illustrates that stronger perturbation is not automatically associated with better OOD performance. Increasing $\rho$ from $0.1$ to $0.9$ increases $D_F$ from approximately $0.158$ to $0.534$ while mean OOD $mAP_{50}$ decreases from $13.7\%$ to $10.4\%$. We do not interpret this as evidence for a causal relation between $D_F$ and OOD accuracy. It is instead a concrete example showing why perturbation strength and downstream performance should not be treated as interchangeable quantities.

Taken together, these experiments suggest a useful division of labor: the perturbation mechanism provides explicit intervention variables, while the descriptor space provides a measurement layer for recording what those interventions actually do. Whether this measurement layer can eventually be used for source-only selection or optimization remains an open question.

\begin{figure}[h]
    \centering
    \includegraphics[width=\linewidth]{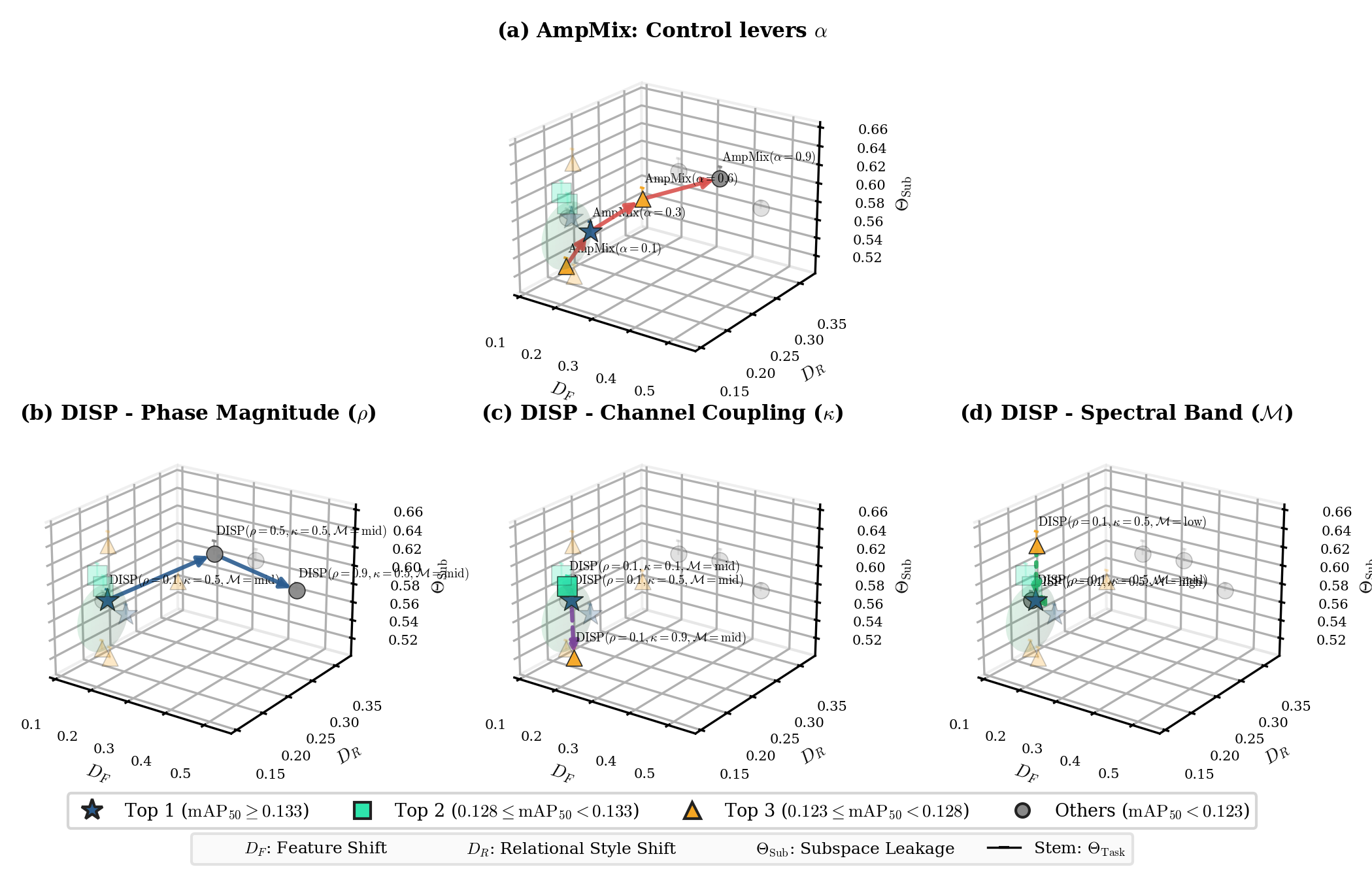}
    \caption{Controlled exploration of the representation descriptor space on Pascal VOC. AmpMix varies along its amplitude-mixing parameter $\alpha$, while DISP generates distinct trajectories by changing phase magnitude $\rho$, channel coupling $\kappa$, and frequency support $\mathcal{M}$. The shaded region is the same descriptive favorable-region guide used in Figure~\ref{fig:sweetspot}.}
    \label{fig:navigation}
\end{figure}

\section{Discussion: What Does This View Open Up?}
\label{sec:discussion}

The main outcome of this work is not a claim that one perturbation is universally superior, but that feature perturbation may benefit from being treated as an object of representation-level analysis. Our measurements provide several preliminary indications of this view: perturbation mechanisms exhibit distinct descriptor profiles even when applied at the same network locations; larger feature displacement does not consistently imply stronger OOD performance; relatively strong configurations tend to occupy a bounded region of the descriptor space; and DISP shows that a simple perturbation operator can expose intervention variables whose effects can be measured rather than inferred only from final accuracy. These observations motivate several questions that extend beyond the current numerical margin of DISP over the strongest baseline. First, does a reproducible favorable region exist across datasets, backbones, detectors, and domain shifts? A stronger evaluation should test whether high-performing perturbations consistently occupy similar regions after suitable normalization. Second, can such a region be identified without target data? Our current analysis uses target performance to characterize favorable configurations and therefore does not yet provide a source-only selection rule. Future SSDG protocols could instead exploit source-domain stability, cross-augmentation consistency, simulated validation domains, or related source-only signals. Third, which descriptors actually matter? Our four descriptors capture magnitude, relational change, spatial allocation, and subspace deviation, but they are neither unique nor necessarily complete; alternatives include CKA-like similarity, covariance geometry, spatial relations, class-conditional distances, attention structure, and task-gradient information. Finally, can descriptor space enable controllable perturbation design? DISP provides an initial step, but future work could formulate perturbation selection as a source-only optimization problem that targets a desired descriptor region while preserving source-task performance. In this view, feature perturbation could move beyond a collection of independent heuristics toward a more systematic design problem. More broadly, descriptor space may offer a useful intermediate vocabulary between intervention mechanism and task performance, provided that its geometry proves stable and informative across broader settings.

\section{Conclusion}
\label{sec:conclusion}

We revisited feature perturbation in SSDG from a representation-centric perspective. Rather than treating perturbation strength as the primary object of study, we introduced a compact descriptor space capturing feature displacement, relational change, task-region allocation, and subspace deviation. Across our experiments, perturbation mechanisms exhibit distinct descriptor profiles, with relatively stronger OOD-performing configurations tending to occupy a bounded portion of this space. DISP provides a simple case study of this perspective by perturbing Fourier phase while explicitly controlling magnitude, channel coupling, and frequency support. This highlights the distinction between \emph{intervention variables} and the \emph{emergent representation descriptors} they produce. The main contribution is therefore a change in viewpoint:

$$
\boxed{\text{Measure}\;\rightarrow\;\text{Characterize}\;\rightarrow\;\text{Question}\;\rightarrow\;\text{Explore}.}
$$

Rather than defining a universal optimal region, this work opens the question of whether such representation-space structure can be identified from source data and used to make feature perturbation more interpretable, controllable, and systematic.

\section*{AI Disclosure}
In this work, we used generative AI tools (large language models) for research execution and methodology refinement, specifically to critically evaluate our scientific rationale, challenge methodological assumptions, and propose refinements to our experimental design. Additionally, we used AI tools to assist in implementing software code for our experimental pipelines and validation tests. We also used generative AI tools to edit and rephrase the manuscript text to improve overall readability, clarity, and language precision. All AI-assisted contributions were rigorously verified by the authors: LLM-generated code was thoroughly inspected, tested, and validated for execution correctness, and all conceptual feedback was critically evaluated through manual literature and theoretical review. We take full responsibility for the final content of this work, including all text, scientific claims, code artifacts, and experimental results.

\bibliography{iclr2027_conference}
\bibliographystyle{iclr2027_conference}

\clearpage
\appendix

\section{DISP Algorithm}
\label{app:algorithm}

Algorithm~\ref{alg:disp} gives the forward pass. The implementation uses a real-valued FFT representation; the non-redundant coefficients determine the corresponding Hermitian-symmetric spectrum for reconstruction. The phase offset is normalized to unit RMS inside the selected band before scaling by $\rho$, and the module is disabled at inference.

\begin{algorithm}[H]
\caption{DISP forward pass}
\label{alg:disp}
\begin{algorithmic}[1]
\Require Feature map $F\in\mathbb{R}^{B\times C\times H\times W}$; $\rho\ge0$; $\kappa\in[0,1]$; frequency band $\mathcal{M}$; probability $p$
\If{not training or $u\sim\mathcal{U}(0,1)\ge p$}
    \State \Return $F$
\EndIf
\State $X\gets\mathcal{F}_2(F)$; $A\gets|X|$; $\varphi\gets\angle X$
\State Sample $z^{\mathrm{shared}}\sim\mathcal{N}(0,1)$ and $z^{\mathrm{indep}}\sim\mathcal{N}(0,1)$
\State $\delta\gets\kappa z^{\mathrm{shared}}+\sqrt{1-\kappa^2}z^{\mathrm{indep}}$
\State Construct radial mask $M_{\mathcal M}$ and set the DC component to zero
\State $\delta\gets\delta\odot M_{\mathcal M}$
\State Normalize $\delta$ to unit RMS on the selected band
\State $\delta\gets\rho\delta$
\State $\varphi'\gets\varphi+\delta$
\State $F'\gets\mathcal{F}_2^{-1}(A\odot e^{i\varphi'})$
\State \Return $F'$
\end{algorithmic}
\end{algorithm}

The module is inserted as a drop-in replacement inside the standard convolution--normalization--activation block, with the perturbation applied between normalization and activation. It is bypassed during inference.

\section{Mathematical Properties}
\label{app:proofs}

The analytical statements below use full-spectrum notation. The implementation stores the non-redundant half-spectrum and reconstructs the Hermitian counterpart needed for real-valued outputs.

\begin{proposition}[Energy invariance]
\label{prop:energy}
DISP preserves the spectral amplitude at every channel and frequency and therefore satisfies
\[
\|F'\|_F=\|F\|_F.
\]
\end{proposition}

\begin{proof}
Under orthonormal Fourier normalization, Parseval's identity gives $\|F\|_F=\|X\|_F$ and $\|F'\|_F=\|A\odot e^{i\varphi'}\|_F$. Since $|A_ce^{i\varphi'_c}|=A_c=|X_c|$, the two spectral norms are equal.
\end{proof}

\begin{corollary}[Bounded continuous feature displacement]
\label{cor:df_bound}
For DISP,
\[
0\leq D_F\leq2,
\qquad
\rho\rightarrow0\Longrightarrow D_F\rightarrow0.
\]
\end{corollary}

\begin{proof}
The bound follows from the triangle inequality and Proposition~\ref{prop:energy}. For the limit, $\varphi'-\varphi=\rho M_{\mathcal M}\odot\delta\to0$ for fixed $\delta$, and continuity of the inverse Fourier transform gives $F'\to F$.
\end{proof}

\begin{lemma}[Gram matrix in the frequency domain]
\label{lemma:gram_freq}
For the channel Gram matrix in Section~\ref{sec:framework},
\[
R(F)_{cc'}=\frac{1}{HW}\sum_{u,v}A_cA_{c'}\cos(\varphi_c-\varphi_{c'}).
\]
\end{lemma}

\begin{proof}
Apply Parseval to the channel-wise inner product and substitute $X_c=A_ce^{i\varphi_c}$.
\end{proof}

\begin{proposition}[Relational invariance at full coupling]
\label{prop:gram_invariance}
Under the full-spectrum/Hermitian reconstruction described above, at $\kappa=1$,
\[
R(F')=R(F),\qquad D_R=0.
\]
\end{proposition}

\begin{proof}
At $\kappa=1$, every channel receives the same phase offset. Therefore $\varphi'_c-\varphi'_{c'}=\varphi_c-\varphi_{c'}$ for every channel pair, while amplitudes remain unchanged. Lemma~\ref{lemma:gram_freq} then gives $R(F')=R(F)$.
\end{proof}

For $0<\kappa<1$, the pairwise phase difference contains the channel-specific term
\[
\rho M_{\mathcal M}\sqrt{1-\kappa^2}
\left(z_c^{\mathrm{indep}}-z_{c'}^{\mathrm{indep}}\right),
\]
so the effect on $D_R$ depends on the existing representation and is not assumed to be monotonic. Likewise, the influence of $\mathcal M$ on the directional descriptors is treated as an empirical question.

\section{Geometric Measurement Protocol}
\label{app:geometry}

\paragraph{Paired features.}
For a fixed source image, clean and perturbed features are obtained from otherwise identical forward passes with the perturbation disabled and enabled, respectively. The detector is kept in evaluation mode during profiling so that batch-normalization statistics remain fixed. For the Baseline, the perturbation is disabled in both passes, yielding $F'=F$ and therefore $D_F=D_R=0$.

\paragraph{Evaluation split.}
Geometric descriptors are computed on the source-domain evaluation split at the $P3/8$ representation, using batches of 16. The same protocol is applied to the source domains of both benchmarks.

\paragraph{Per-image computation.}
With clean/perturbed features $F_c,F_p$ and ground-truth boxes, Algorithm~\ref{alg:geometry} implements the four definitions in Section~\ref{sec:framework}. We use $k_{\mathrm{ratio}}=0.2$ throughout.

\begin{algorithm}[H]
\caption{Per-image representation descriptors}
\label{alg:geometry}
\begin{algorithmic}[1]
\Require $F_c,F_p\in\mathbb{R}^{C\times H\times W}$; ground-truth boxes $\mathcal B$; stride $s$; $k_{\mathrm{ratio}}$
\State $\Delta\gets F_p-F_c$
\State $D_F\gets\|\Delta\|_F/(\|F_c\|_F+\varepsilon)$
\State $R_c,R_p\gets\frac{1}{HW}F_{c,(2)}F_{c,(2)}^\top,\frac{1}{HW}F_{p,(2)}F_{p,(2)}^\top$
\State $D_R\gets\|R_p-R_c\|_F/(\|R_c\|_F+\varepsilon)$
\State Project the union of boxes to the feature grid to obtain $M$; $\bar M\gets1-M$
\State $D_{\mathrm{obj}}\gets\|\Delta\odot M\|_F/(\|F_c\odot M\|_F+\varepsilon)$
\State $D_{\mathrm{bg}}\gets\|\Delta\odot\bar M\|_F/(\|F_c\odot\bar M\|_F+\varepsilon)$
\State $\Theta_{\mathrm{Task}}\gets D_{\mathrm{bg}}/(D_{\mathrm{obj}}+\varepsilon)$
\State Compute $V_K$ from the top-$K$ right singular vectors of $F_{c,(2)}^\top$, $K=\lceil k_{\mathrm{ratio}}C\rceil$
\State $E_{\mathrm{tot}}\gets\|\Delta_{(2)}\|_F^2$; $E_{\parallel}\gets\|\Delta_{(2)}^\top V_K\|_F^2$
\State $\Theta_{\mathrm{Subspace}}\gets(E_{\mathrm{tot}}-E_{\parallel})/E_{\mathrm{tot}}$
\State \Return $D_F,D_R,\Theta_{\mathrm{Task}},\Theta_{\mathrm{Subspace}}$
\end{algorithmic}
\end{algorithm}

Images without annotated objects are excluded from aggregate $\Theta_{\mathrm{Task}}$ computation. The two directional descriptors are also excluded when $D_F$ falls below a fixed numerical tolerance. Reported geometric values are means over the remaining valid images.

\section{Datasets and Domain Splits}
\label{app:datasets}

\begin{figure*}[h]
    \centering
    \includegraphics[width=\textwidth]{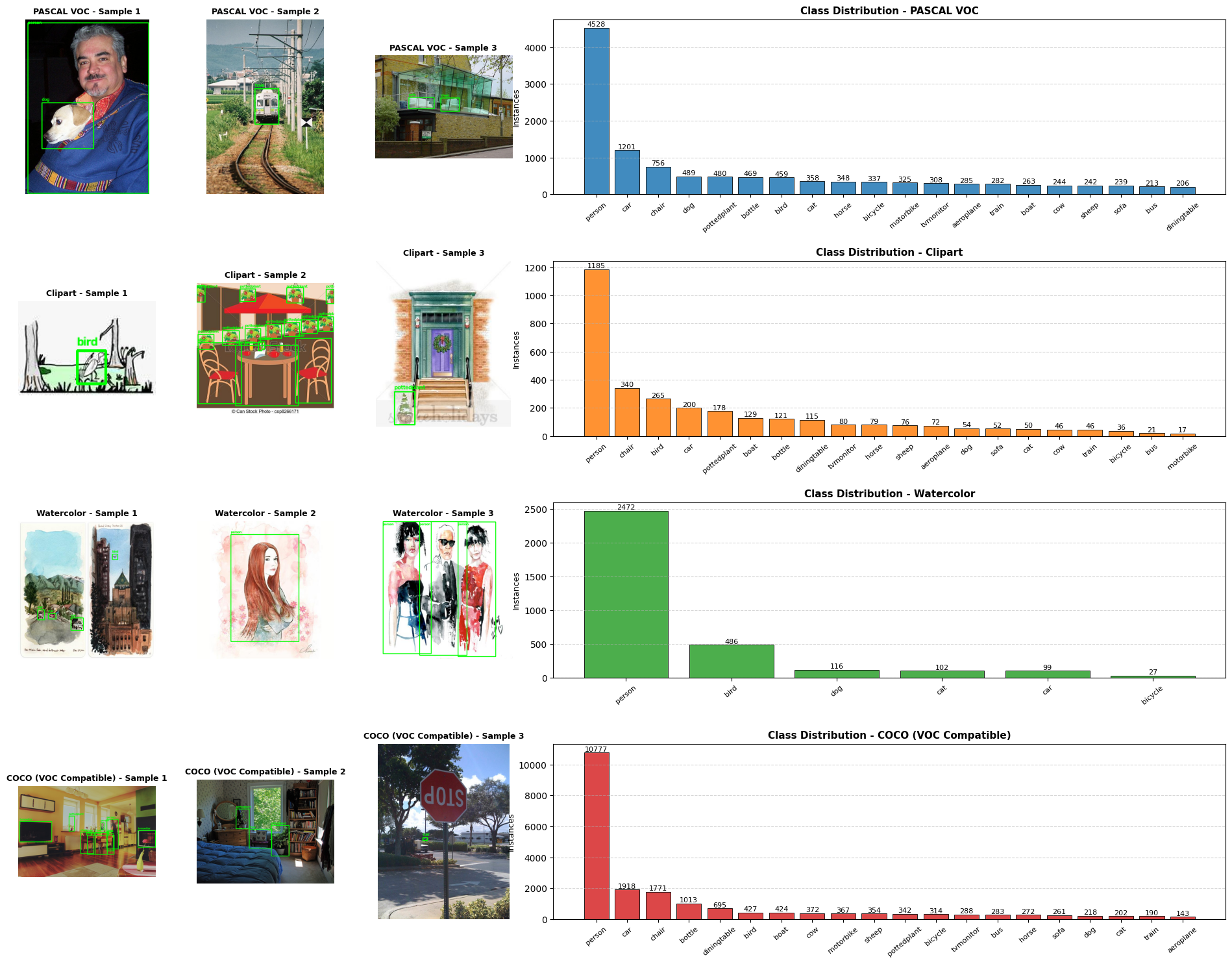}
    \caption{\textbf{Cross-Domain Object Detection Benchmark and Category Distribution.} 
    Visual samples with bounding box annotations (left) alongside class frequency distributions (right) across four benchmark domains: \textit{PASCAL VOC} (source/real), \textit{Clipart} (vector graphics), \textit{Watercolor} (artistic style), and \textit{COCO} (VOC-compatible split). The figure highlights both the substantial \textit{style/visual shift} across domains and the intrinsic \textit{category imbalance} within each dataset, illustrating the challenges of cross-domain object detection.}
    \label{fig:benchmark_overview}
\end{figure*}

\paragraph{Style-shift benchmark: Pascal VOC $\rightarrow$ Clipart, Watercolor, COCO.}
The source domain is Pascal VOC~2007, comprising the standard 20-class
label space
\[
\{\text{aeroplane, bicycle, bird, boat, bottle, bus, car, cat, chair, cow,}
\]
\[
\text{diningtable, dog, horse, motorbike, person, pottedplant, sheep, sofa, train, tvmonitor}\},
\]
with 5{,}011 training images and 4{,}952 test images (VOC$_{\text{Orig}}$
in Table~\ref{tab:main_merged_with_source}), used as the sole training and
in-domain evaluation data. Clipart and Watercolor originate from the
cross-domain detection benchmark of \citet{inoue_cross-domain_2018}.
Clipart shares the full 20-class VOC label space and contributes 999 test
images; Watercolor covers only a 6-class subset,
$\{\text{person, bird, dog, cat, car, bicycle}\}$, and contributes 1{,}997
test images, so mean OOD statistics involving Watercolor are computed only
over these six shared classes, not the full 20. COCO$_{\text{Comp}}$
denotes the COCO validation set filtered and remapped to the VOC label
space, contributing 4{,}952 test images. No target-domain data is used
during training in any configuration. Figure~\ref{fig:benchmark_overview}
(``Overview of Visual Style Divergence and Class Imbalance across PASCAL
VOC, Clipart, Watercolor, and COCO'') shows representative images and the
per-class instance distribution for all four domains.

\begin{table}[htbp]
\centering
\caption{Style-shift benchmark: image counts and label space by domain.}
\label{tab:voc_family_stats}
\begin{tabular}{@{}lccc@{}}
\toprule
Domain & Train & Test & Classes \\
\midrule
VOC$_{\text{Orig}}$ (source) & 5{,}011 & 4{,}952 & 20 \\
Clipart (target)             & --      & 999     & 20 \\
Watercolor (target)          & --      & 1{,}997 & 6 (shared subset) \\
COCO$_{\text{Comp}}$ (target) & --     & 4{,}952 & 20 (remapped) \\
\bottomrule
\end{tabular}
\end{table}

\begin{figure} [H]
    \centering
    \includegraphics[width=1\linewidth]{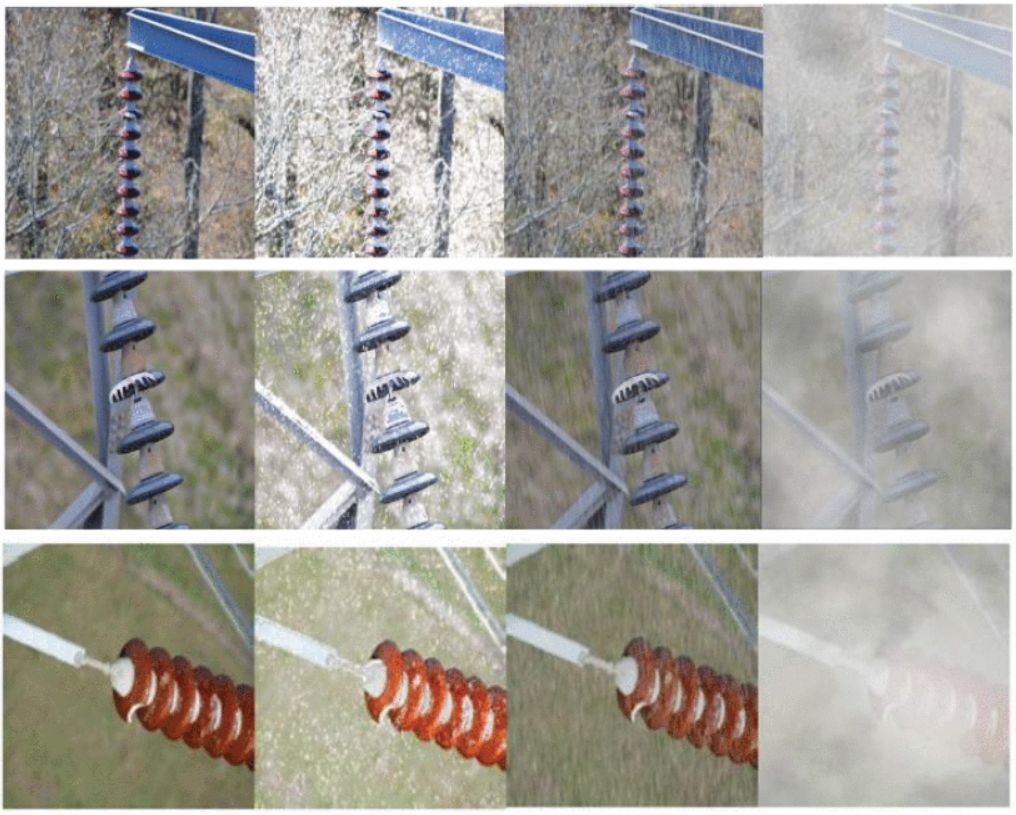}
    \caption{Illustrative Images of Datasets: Sequential column display showing IDID-Original (first column), followed by IDID-Snow, IDID-Rain, and IDID-FOG (last column).\citep{benelmostafa_yolov8-erfd-ap_2025}}
\label{fig:idid_overview}
\end{figure}

\paragraph{Weather-shift benchmark: IDID clear-weather $\rightarrow$ rain, fog, snow.}
IDID~\citep{aitelhaj_weathering_2026} is a UAV power-line inspection
dataset comprising 1{,}600 images, split 70\%/20\%/10\% into
training/test/validation (approximately 1{,}120/320/160 images). 
The source domain (IDID$_{\text{Orig}}$ in
Table~\ref{tab:main_merged_with_source}) consists of clear-weather images
only; models are trained exclusively on this split. A single anomaly class is annotated, representing insulator shells that are either flashover-damaged or broken. The three
target domains, IDID\_Rain, IDID\_Fog, and IDID\_Snow, apply simulated
weather conditions to the corresponding test images, following the
protocol of \citet{benelmostafa_yolov8-erfd-ap_2025}, and are never seen
during training. Figure~\ref{fig:idid_overview} shows representative
images from IDID$_{\text{Orig}}$, IDID\_Snow, IDID\_Rain, and IDID\_Fog
side by side, illustrating the visual severity of each condition.

\begin{table}[htbp]
\centering
\caption{Weather-shift benchmark: image counts and label space.}
\label{tab:idid_stats}
\begin{tabular}{@{}lccc@{}}
\toprule
Domain & Train & Test & Classes \\
\midrule
IDID$_{\text{Orig}}$ (source) & $\sim$1{,}120 & $\sim$320 & 1 \\
IDID\_Rain (target)  & -- & $\sim$320 & 1 \\
IDID\_Fog (target)   & -- & $\sim$320 & 1 \\
IDID\_Snow (target)  & -- & $\sim$320 & 1 \\
\bottomrule
\end{tabular}
\end{table}

\section{Implementation and Training Details}
\label{app:implementation}

All experiments use Ultralytics~v8.4.49 and YOLO26-nano under otherwise identical optimization settings. Pascal VOC experiments use $512\times512$ inputs and 70 epochs; IDID uses $640\times640$ inputs and 100 epochs without pretrained weights. Feature perturbations are applied stochastically with $p=0.35$ and disabled during evaluation. The training runs are tracked with ClearML on a single NVIDIA T4 GPU.

The same perturbation locations are used for all compared methods. For methods whose original formulation uses a different intervention level, the current study adopts a common feature-level implementation at the same backbone locations so that the comparison isolates the behavior of the perturbation mechanism under a shared detector/training pipeline. This should be understood as a level-matched experimental adaptation rather than a claim of exact reproduction of every original training recipe.

\begin{figure}[H]
    \centering
    \begin{subfigure}[b]{0.48\textwidth}
        \centering
        \includegraphics[width=\linewidth]{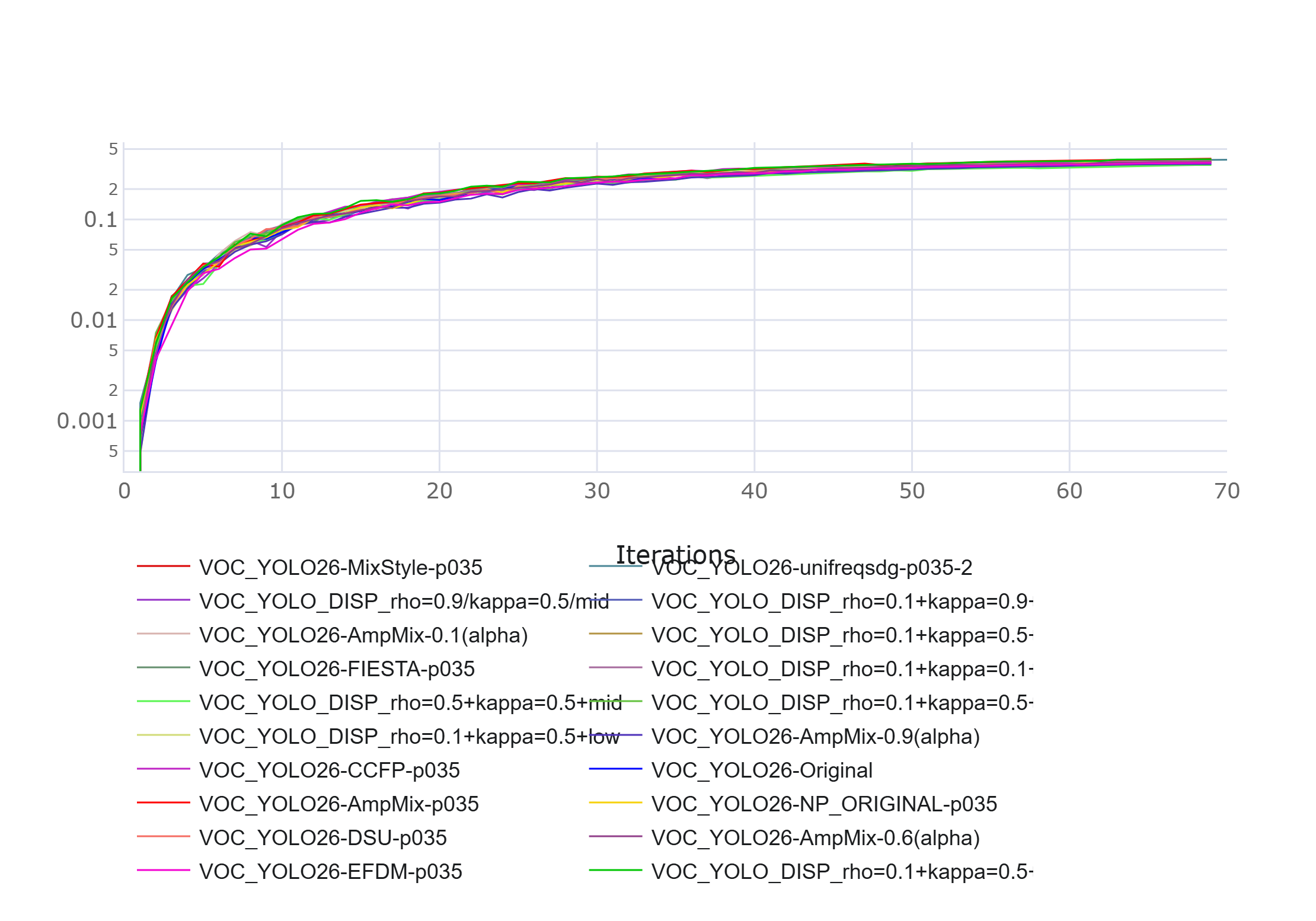}
        \caption{VOC: $mAP_{50}$}
    \end{subfigure}\hfill
    \begin{subfigure}[b]{0.48\textwidth}
        \centering
        \includegraphics[width=\linewidth]{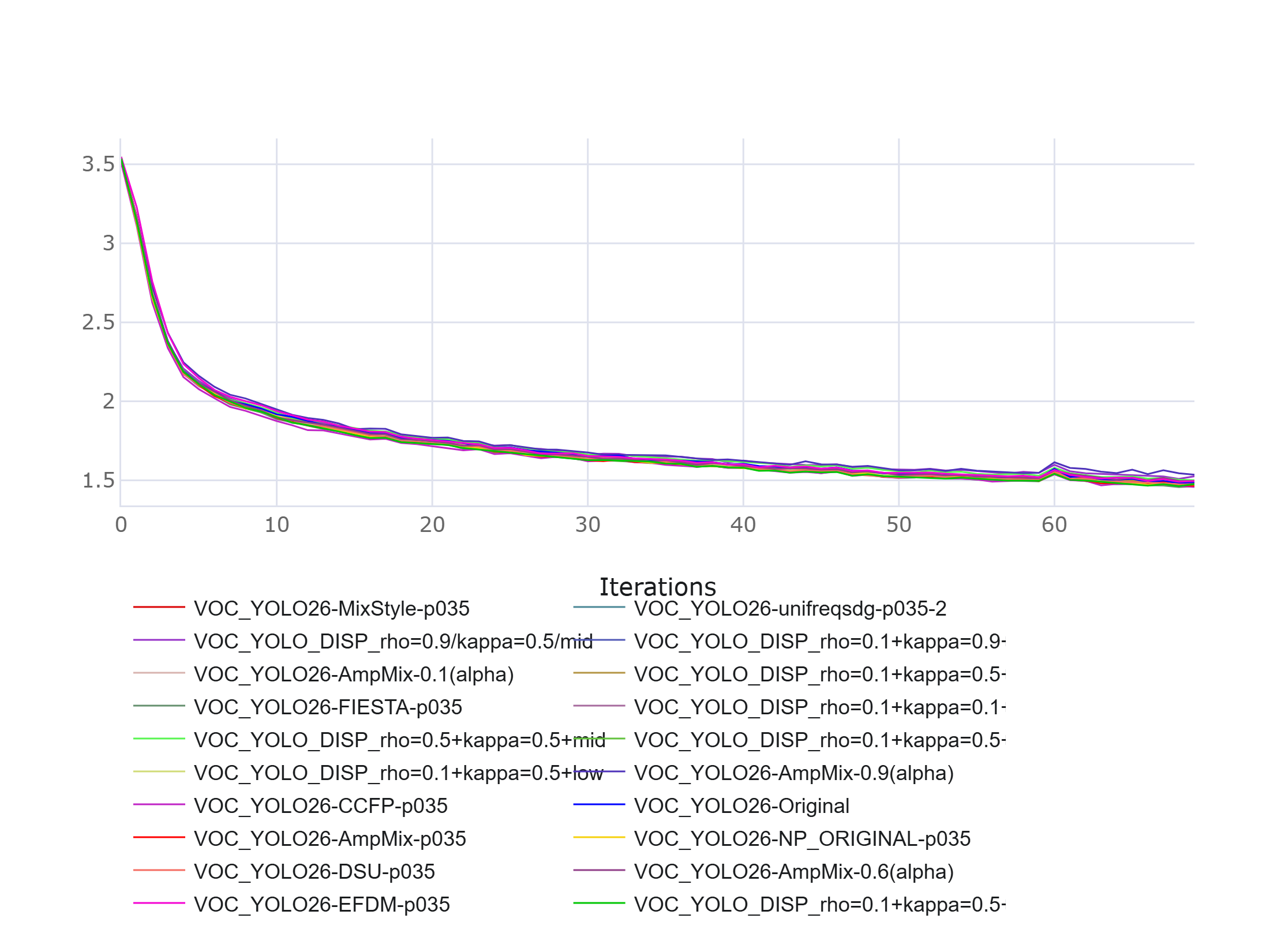}
        \caption{VOC: box loss}
    \end{subfigure}
    \caption{Representative training curves on the VOC benchmark.}
    \label{fig:voc_curves}
\end{figure}

\begin{figure}[H]
    \centering
    \begin{subfigure}[b]{0.48\textwidth}
        \centering
        \includegraphics[width=\linewidth]{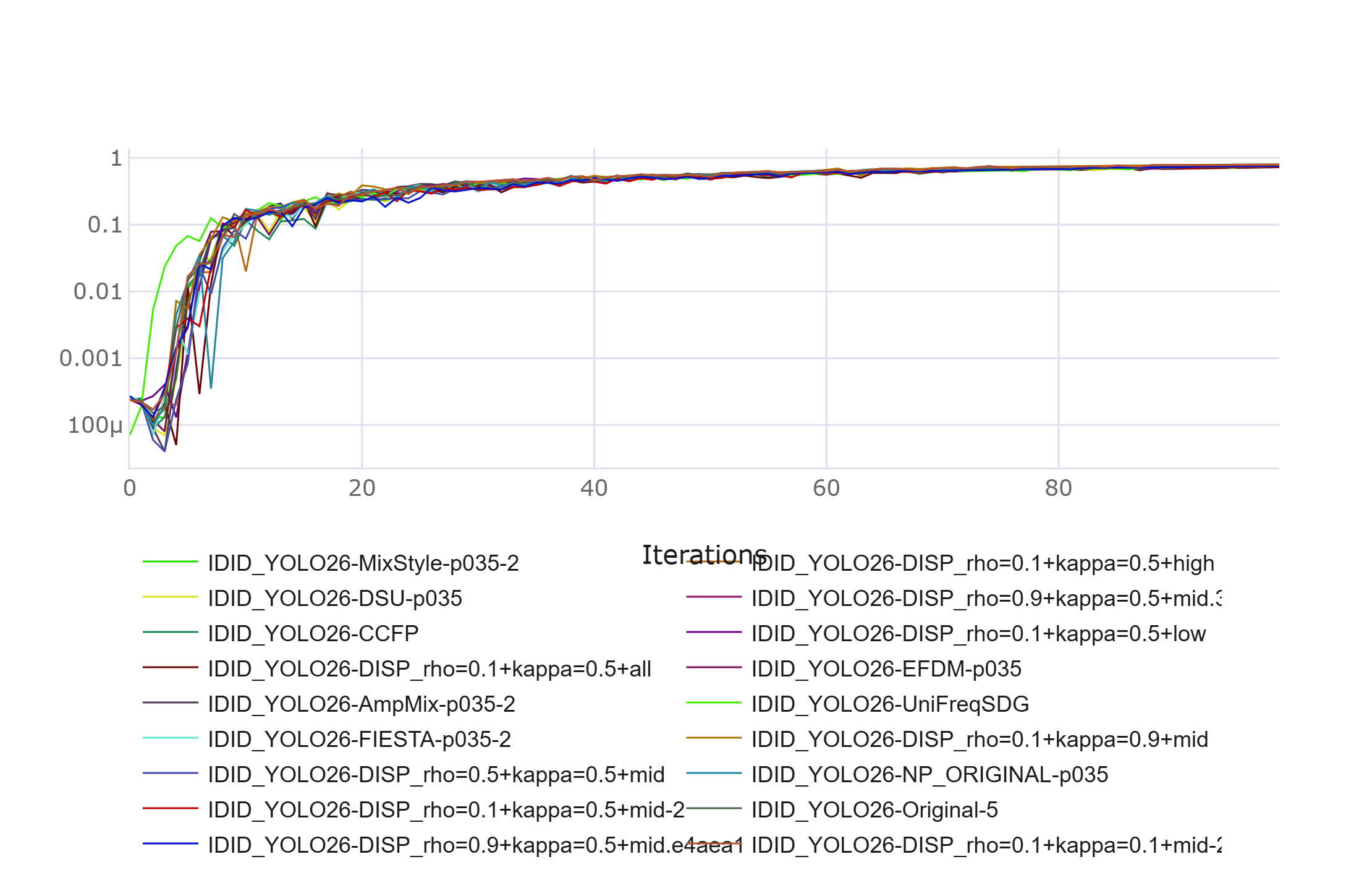}
        \caption{IDID: $mAP_{50}$}
    \end{subfigure}\hfill
    \begin{subfigure}[b]{0.48\textwidth}
        \centering
        \includegraphics[width=\linewidth]{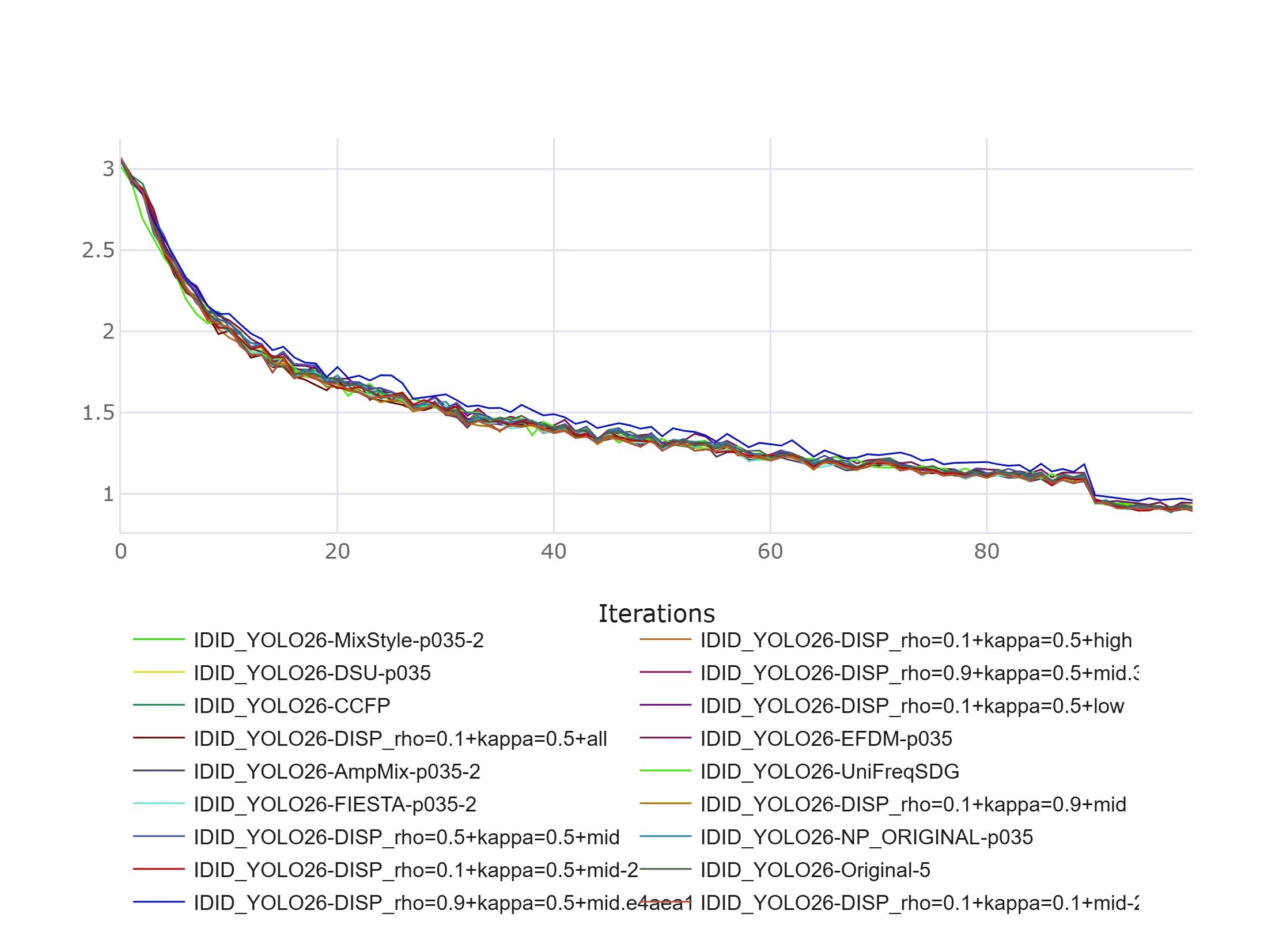}
        \caption{IDID: box loss}
    \end{subfigure}
    \caption{Representative training curves on the IDID benchmark.}
    \label{fig:idid_curves}
\end{figure}

\section{Extended Results}
\label{app:full_results}

\subsection{Controlled Exploration: DISP Sweep}
\label{app:navigation}

\begin{table}[H]
\centering
\small
\caption{DISP parameter sweep on Pascal VOC. Values are $\mathrm{mAP}_{50}$ in percent, followed by the four representation descriptors. The sweep is exploratory and includes target-domain scores.}
\label{tab:disp_ablation}
\resizebox{\textwidth}{!}{%
\begin{tabular}{l|ccccc|cccc}
\toprule
\textbf{Config} & VOC & Clipart & Water. & COCO & Mean & $D_F$ & $D_R$ & $\Theta_{\mathrm{Task}}$ & $\Theta_{\mathrm{Subspace}}$ \\
\midrule
$(0.1,0.5,\mathrm{mid})$ & 27.8 & 6.7 & 10.7 & 9.6 & 13.7 & 0.158 & 0.197 & 1.143 & 0.574 \\
$(0.1,0.1,\mathrm{mid})$ & 26.7 & 5.6 & 9.2 & 10.1 & 12.9 & 0.153 & 0.194 & 1.144 & 0.589 \\
$(0.1,0.9,\mathrm{mid})$ & 24.9 & 5.8 & 9.4 & 9.7 & 12.4 & 0.160 & 0.199 & 1.147 & 0.510 \\
$(0.1,0.5,\mathrm{all})$ & 27.2 & 5.8 & 8.4 & 9.7 & 12.8 & 0.157 & 0.182 & 1.154 & 0.605 \\
$(0.1,0.5,\mathrm{low})$ & 25.8 & 5.0 & 9.7 & 9.4 & 12.5 & 0.188 & 0.182 & 1.180 & 0.640 \\
$(0.1,0.5,\mathrm{high})$ & 25.6 & 4.6 & 6.1 & 8.9 & 11.3 & 0.147 & 0.197 & 1.121 & 0.573 \\
$(0.5,0.5,\mathrm{mid})$ & 22.5 & 4.5 & 5.1 & 8.6 & 10.2 & 0.393 & 0.235 & 1.170 & 0.636 \\
$(0.9,0.5,\mathrm{mid})$ & 23.4 & 5.0 & 5.2 & 8.0 & 10.4 & 0.534 & 0.290 & 1.091 & 0.597 \\
\bottomrule
\end{tabular}}
\end{table}

\begin{table}[H]
\centering
\small
\caption{AmpMix sweep on Pascal VOC. Values are $\mathrm{mAP}_{50}$ in percent, followed by the four representation descriptors.}
\label{tab:ampmix_ablation}
\begin{tabular}{l|ccccc|cccc}
\toprule
$\alpha$ & VOC & Clipart & Water. & COCO & Mean & $D_F$ & $D_R$ & $\Theta_{\mathrm{Task}}$ & $\Theta_{\mathrm{Subspace}}$ \\
\midrule
0.1 & 26.8 & 5.4 & 7.8 & 9.7 & 12.4 & 0.134 & 0.201 & 1.111 & 0.517 \\
0.3 & 27.8 & 6.1 & 9.8 & 9.7 & 13.3 & 0.165 & 0.227 & 1.128 & 0.551 \\
0.6 & 24.8 & 6.6 & 9.1 & 8.7 & 12.3 & 0.233 & 0.280 & 1.153 & 0.578 \\
0.9 & 22.9 & 4.5 & 8.0 & 8.0 & 10.9 & 0.326 & 0.368 & 1.160 & 0.585 \\
\bottomrule
\end{tabular}
\end{table}

The DISP sweep illustrates a clear magnitude effect in the current runs: increasing $\rho$ from $0.1$ to $0.9$ raises $D_F$ from $0.158$ to $0.534$ in the mid-band setting, while mean OOD $mAP_{50}$ decreases from $13.7\%$ to $10.4\%$. Changes in $\kappa$ at fixed $\rho$ alter the descriptor profile without a simple monotonic trend in $D_R$. Frequency selection also changes the profile, supporting the use of $\mathcal{M}$ as an intervention variable. These results should be read as sensitivity observations, not as evidence that any one direction in descriptor space is intrinsically optimal.

\subsection{Computational Analysis}
\label{app:computational_analysis}

DISP adds no trainable parameters and is bypassed during inference. The reported model size remains approximately $2.51$M parameters. The inference measurements below reflect the disabled-perturbation evaluation path and therefore should not be interpreted as training-time Fourier overhead.

\begin{table}[H]
\centering
\small
\caption{Model size and representative inference measurements.}
\label{tab:computational_summary}
\begin{tabular}{lcc}
\toprule
Method & Params (M) & Avg. Speed (ms) \\
\midrule
Baseline & 2.51 & 2.46 $\pm$ 0.21 \\
AmpMix & 2.51 & 2.49 $\pm$ 0.20 \\
EFDM & 2.51 & 2.36 $\pm$ 0.14 \\
FIESTA & 2.51 & 2.32 $\pm$ 0.16 \\
DSU & 2.51 & 2.40 $\pm$ 0.18 \\
MixStyle & 2.51 & 2.38 $\pm$ 0.14 \\
NP$_{\mathrm{Original}}$ & 2.51 & 2.34 $\pm$ 0.15 \\
CCFP & 2.51 & 2.29 $\pm$ 0.15 \\
UniFreqSDG & 2.52 & 2.28 $\pm$ 0.13 \\
\textbf{DISP} & \textbf{2.51} & 2.66 $\pm$ 0.18 \\
\bottomrule
\end{tabular}
\end{table}

\subsection{Qualitative Feature-Map Analysis}
\label{app:feature_analysis}

The feature-map visualization complements the descriptor measurements. EFDM is a clear geometric outlier on VOC, with $D_F\approx0.43$ and $D_R\approx0.72$. The corresponding shallow-layer activations show more substantial changes in localized response patterns than methods occupying the central part of the descriptor space. We treat this only as a qualitative illustration of the measured displacement, not as direct evidence of semantic distortion or causality.

DISP remains within the moderate-displacement range observed among the compared competitive configurations. Its representative feature responses change while retaining recognizable object and boundary structure in the displayed examples.

\begin{figure}[H]
    \centering
    \includegraphics[width=\linewidth]{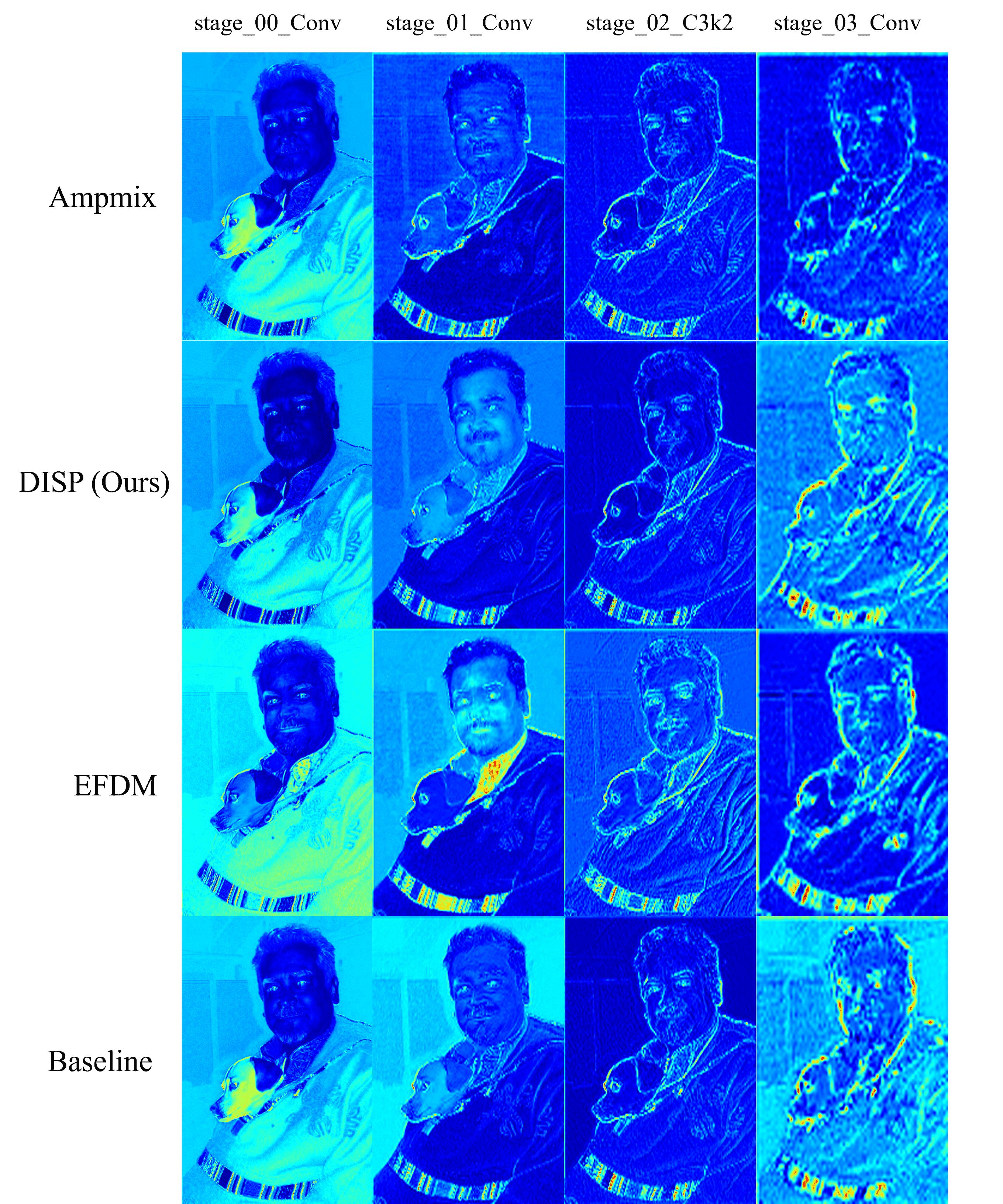}
    \caption{Representative shallow-layer feature-map responses. EFDM shows substantially larger measured displacement, while DISP exhibits more moderate changes consistent with its descriptor profile.}
    \label{fig:feature_maps}
\end{figure}

\end{document}